\documentclass[11pt,a4paper]{article}
\usepackage{amssymb}
\usepackage{graphicx}

\usepackage{url}
\usepackage[T1]{fontenc}
\usepackage[utf8]{inputenc}
\usepackage{CJKutf8}
\usepackage{amsmath}
\usepackage{algorithm}

\usepackage{multirow}
\usepackage{algorithm}
\usepackage{algorithmic}
\usepackage{tabularx}   
\usepackage{makecell}   
\usepackage{caption}
\usepackage{amsthm}
\usepackage{booktabs}
\usepackage{algorithmic}
\usepackage[switch]{lineno}
\usepackage{fontawesome}  
\usepackage{makecell} 

\title{FAGC: Feature Augmentation on Geodesic Curve in the Pre-Shape Space}

\author{Yuexing Han*, Gan Hu, Guanxin Wan and Bing Wang*}

\begin{document}
\begin{CJK}{UTF8}{gbsn}
\maketitle

\begin{abstract}
Due to the constraints on model performance imposed by the size of the training data, data augmentation has become an essential technique in deep learning. However, most existing data augmentation methods are affected by information loss and perform poorly in small-sample scenarios, which limits their application. To overcome the limitation, we propose a Feature Augmentation method on Geodesic Curve in the pre-shape space, called the FAGC. First, a pre-trained neural network model is employed to extract features from the input images. Then, the image features as a vector is projected into the pre-shape space by removing its position and scale information. In the pre-shape space, an optimal  Geodesic curve is constructed to fit the feature vectors. Finally, new feature vectors are generated for model learning by interpolating along the constructed Geodesic curve. We conducted extensive experiments to demonstrate the effectiveness and versatility of the FAGC. The results demonstrate that applying the FAGC to deep learning or machine learning methods can significantly improve their performance in small-sample tasks.
\end{abstract}

Keywords: Feature augmentation, Shape space, Pre-shape space, The Geodesic curve

\section{INTRODUCTION}
\label{sec:introduction}
As a powerful artificial intelligence technology, deep learning has made revolutionary breakthroughs in computer vision, natural language processing, speech recognition, and other fields \cite{lecun2015deep,otter2020survey,malik2021automatic}. A well-performing deep learning model typically requires large-scale, high-quality labeled data for training, but it is often not available in reality. The small-sample problem has become a major factor impeding the broader application of deep learning methods in certain real-world scenarios. For instance, in the field of materials science, acquiring a single SEM image of a copper alloy is often costly, potentially amounting to several hundred dollars. The key to addressing the small-sample problem lies in providing sufficient knowledge representation from limited samples, thereby enhancing its generalization capability. 

Some novel approaches addressing the small-sample challenge have recently emerged, showing significant performance gains across multiple benchmarks. 
Transfer learning is a mainstream approach in deep learning to address the problem of data scarcity \cite{hossen2025transfer}. Its core idea is to transfer knowledge obtained from large-scale datasets to the training process of the target dataset, mitigating the lack of information. In practical applications, transfer learning most commonly follows a two-stage training paradigm, i.e., pre-training and fine-tuning. The transfer model first learns generic feature extraction capabilities on a large-scale dataset, and is subsequently fine-tuned on the target dataset to adapt to specific tasks. However, the methodology necessitates a large-scale source dataset relevant to the target domain, which often unmet in practical scenarios. Crucially, transfer performance is highly dependent on the inter-dataset relevance. Leveraging transfer learning strategies, meta-learning-based Few-Shot Learning (FSL) have been proposed to address the small-sample problem \cite{zhang2025few}. The FSL first acquires a similarity function by training on a large-scale base dataset rich in category and sample information, developing the model's ability to adapt to novel categories. During inference, the FSL makes final predictions by computing pairwise similarity measures between query set instances and support set prototypes. Although the FSL solutions are generally applicable and can adapt to most few-shot target datasets, they often prove less efficient when the target task is well-defined. 
By contrast, data augmentation offers a more straightforward solution to small-sample problems through artificial expansion of limited datasets \cite{su2024enhanced}.

While conventional augmentation strategies (e.g., geometric transformations or color jittering) offer baseline improvements, they often fail to capture complex data distributions, especially in small-sample scenarios. With the advancement of deep learning, data augmentation methodologies have evolved along two main directions by leveraging the powerful representational capabilities of large-scale models. One approach is to leverage deep generative techniques such as GAN \cite{goodfellow2014generative}, VAE \cite{kingma2013auto}, and Diffusion \cite{ho2020denoising} models to synthesize virtual samples that conform to the real data distribution for training. While these virtual data enhance sample diversity, developing effective generative models under small-sample constraints remains a significant challenge. In addition, in some fields with high security requirements, such as medical applications, the generated virtual data lacks sufficient reliability. Another approach is to directly use large models to extract image features, and then data augmentation is achieved at the feature level. In Ref \cite{terrance2017dataset}, the probability of encountering real samples in the feature space is higher than that in the input space when traversing along the manifold. Despite boosting downstream task performance, current feature augmentation techniques remain perform poorly in small-sample settings. 

To compensate for this deficiency, we introduce the shape space theory to propose Feature Augmentation on Geodesic Curve (FAGC) in pre-shape space, delivering effective feature augmentation under data scarcity. The shape space theory \cite{kendall1984shape,kendall2009shape,kendall1977diffusion}, as a kind of manifold theory, mainly focuses on the geometric attributes and structures of the data. The theory has been widely used in image processing due to its remarkable interpretability of data and insensitivity to limited datasets \cite{zhang2003object,glover2006robust,kume2007shape,kilian2007geometric}. As demonstrated in \cite{hu2025projection}, the pre-shape space exhibits inherent robustness to small-sample data and enables seamless integration with deep learning models. Furthermore, as indicated in \cite{han2010recognition}, the Geodesic curves in the pre-shape space contain rich geometric transformation information that can improve object recognition accuracy. Building upon the current research foundation, the FAGC provides a comprehensive manifold feature augmentation framework for machine learning pipelines. The FAGC first represents image data as feature vectors within the pre-shape space, involving processes such as feature extraction, modeling, and projection. The core of the FAGC lies in its designed algorithm, which constructs a optimal Geodesic curve in the pre-shape space that fits multiple initial feature vectors. The curve minimizes the sum of Geodesic distances to all initial feature vectors under the metric of the pre-shape space. Then, new feature vectors are sampled along the curve to achieve feature augmentation. Finally, to effectively utilize both the augmented data and the real data during downstream task training, a corresponding loss function construction method is formulated.

\begin{figure*}[t]  
     \centering
    \includegraphics[width=\textwidth]{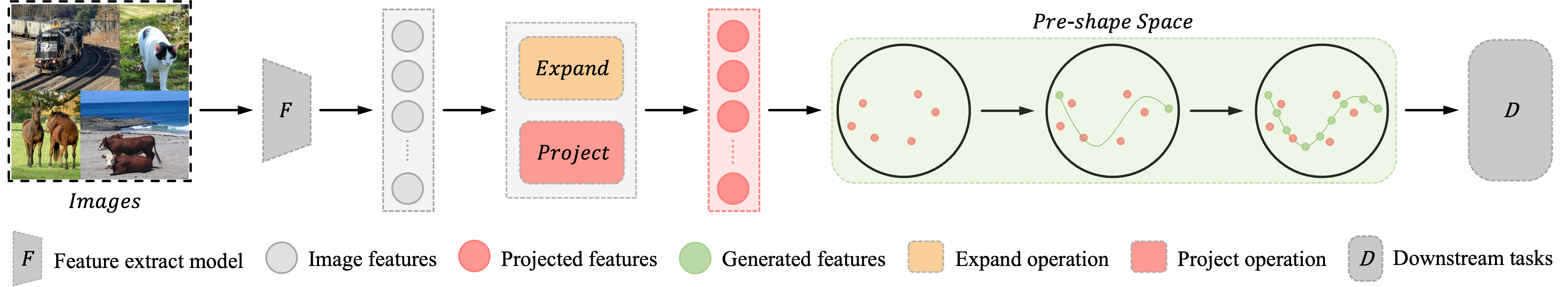}
    \caption{\rmfamily{An overview of the FAGC method.}}
    \label{fig:overview}
\end{figure*}

Figure~\ref{fig:overview} provides a concise overview of our framework. Our main contributions are summarized as follows:
\begin{itemize}
    \item A feature augmentation method called the FAGC is proposed. The FAGC achieves efficient feature augmentation under small-sample conditions by constructing the optimized Geodesic curves in the pre-shape space. 
    \item The FAGC can be seamlessly integrated into the machine learning methods or deep learning methods, demonstrating its strong performance in extensive small-sample scenarios, particularly when data is extremely scarce.
    \item Multiple ablation studies bridge the theoretical foundation of the FAGC to its practical applications. The visual experiments further validate the safety and reliability of the data augmented by the FAGC, expanding its applicability scope.
\end{itemize}

\section{Related Work}
\subsection{Pixel-level augmentation}
The effectiveness of data augmentation was first demonstrated through simple pixel-level transformations, such as horizontal flipping, color enhancement, and random cropping \cite{shorten2019survey}. These pixel-level methods have some disadvantages. First, the distribution of the perturbation images ought to be close to the distribution of the original images. Furthermore, some perturbation image methods, such as translation and rotation, suffer from the padding effect. After the operation is completed, part of the image is moved out of the boundary and lost. Thus, some interpolation methods select one of the four pixels around the blank position and use it to fill the blank area \cite{rukundo2012nearest}. These processing methods weaken the model's attention to the critical regions and affect the model's final performance. Recent studies have placed greater emphasis on sophisticated methods, such as Random Erasure \cite{2017Random}, Mixup \cite{zhang2017mixup}, Cutout \cite{devries2017improved}, RandAugment \cite{cubuk2020randaugment}, etc. These methods effectively increase the amount of training data, but the improvement in diversity is limited.
With the advancement of large-scale models, data-driven pixel-level image reconstruction has been leveraged for data augmentation. 
For example, AutoMix \cite{2022AutoMix} proposes an automated mixing framework that enables models to generate mixed samples conducive to classification performance. 
Another reconstruction method is generative modeling, which generates the realistic image from the real image, such as GAN \cite{goodfellow2014generative}, VAE \cite{kingma2013auto}, and Diffusion \cite{ho2020denoising}. These models have been widely used due to their excellent generative capabilities. However, the quality of generative models still heavily depends on the amount of data, often resulting in poor image quality when the training data is limited.

\subsection{Feature-level augmentation}
Feature-level augmentation primarily refers to the method of transforming and augmenting data in the feature space. 
Extracting distinctive features from images is a prerequisite for feature augmentation. Early methods typically relied on manually designed feature operators, such as SIFT \cite{lowe2004distinctive}, HOG \cite{dalal2005histograms}, and LBP \cite{kylberg2011virus}, which were often less effective. With the advancement of deep learning, deep models have been employed to replace these manual approaches for extracting high-quality features. Consequently, feature augmentation has gained significant research attention relying on deep learning techniques.
For example, the stochastic feature-augmented SFA algorithm expands data by randomly applying noise information to the output of the intermediate layer during model training \cite{li2021simple}. 
In \cite{li2021feature}, an implicit feature augmentation method called moment exchange is proposed. The method encourages the model to replace the current moment information with the moment information of latent features at another time.
There are also some feature augmentation methods that are improvements upon existing input-space data augmentation techniques. As in \cite{liu2024feature}, SMOTE is applied at the feature level, enhancing the accuracy of imbalance fault diagnosis in gas turbines.
Similarly in \cite{verma2019manifold}, feature-level Mixup techniques, when applied to intermediate neural network layers, have successfully enhanced model performance.
Although existing feature augmentation methods have demonstrated effectiveness, further improving model performance in small-sample scenarios remains a significant challenge.
In this work, we leverage the shape space theory, which has been proven effective for small-sample scenarios \cite{hu2025projection}, to enable efficient feature augmentation in the manifold space.

\subsection{The shape space theory}
\begin{figure}[htbp]
    \centering
    \includegraphics[scale=0.5]{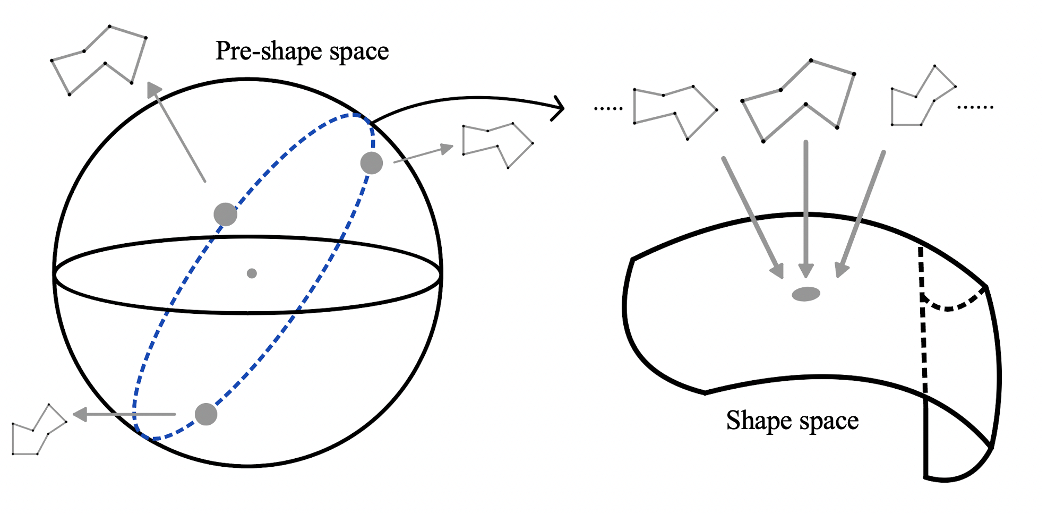}
    \caption{\rmfamily{A pre-shape space and a shape space. A shape with different positions, scales, and orientations all lie on an orbit in the pre-shape space, while this orbit is a point in the shape space.}}
    \label{figure2}
\end{figure}

The shape space theory, firstly introduced by Kendall, is used to describe an object and all its equivalent variations in a non-Euclidean space \cite{kendall1984shape}. In the shape space, the position, scale, and orientation of a shape can be neglected. The shape space theory is often used in object recognition scenarios, where the distance between objects in shape space determines the outcome of object recognition. The feature representing the shapes of the objects are first projected into the pre-shape space, i.e., all the shapes of the features of the $p$ coordinate points are embedded in a unit hyper-sphere \cite{kendall2009shape,small2012statistical}, denoted as $S_{*}^{2p-3}$. Any shape is a point or vector on this hyper-sphere, and all changes in the shape, i.e., position, scale scaling, and 2D rotation, result in a new shape that lies on a great circle in $S_{*}^{2p-3}$, denoted $O(v)$, where $v$ denotes that the shape is on $S_{*}^{2p-3}$. The set of all shapes in the unit hyper-sphere $S_{*}^{2p-3}$ forms the orbit space $\sum_{2}^p$, which is a shape space described as follows:
\begin{equation}
    \sum{\substack{p\\[0.5em]2}} = \{O(v):v\in S_{*}^{2p-3}\}. \tag{1} \label{Eq.1}
\end{equation}
An example of pre-shape space and shape space is shown in Figure \ref{figure2}. Pre-shapes of the same polygon at different locations, scales, and orientations are all on the same great circle of $S_{*}^{2p-3}$, where the great circle serves as a point or vector in the shape space. Rather than the Euclidean distance, the Geodesic distance is the shortest distance between two points in the pre-shape space $S_{*}^{2p-3}$. Assuming that $v_1$ and $v_2$ are two features in the pre-shape space, the Geodesic distance between $v_1$ and $v_2$ is defined as \cite{han2010recognition}:
\begin{equation}
    d(v_{1},v_{2})=\cos^{-1}(\langle v_{1},v_{2} \rangle), \tag{2} \label{Eq.2}
\end{equation}
where $(\langle v_{1},v_{2} \rangle)$ denotes the inner product between $v_1$ and $v_2$. The distance between two shapes in the shape space $\sum_{2}^{p}$ can be defined as follows \cite{kendall2009shape,small1996statistical}:
\begin{equation}
    d_{p}[O(v_{1}),O(v_{2})]=\text{inf}[\cos^{-1}(\langle \alpha,\beta \rangle):\alpha\in O(v_{1}),\beta\in O(v_{2})]. \tag{3} \label{Eq.3}
\end{equation}
    However, the $d_p$ calculation in the formula is very complicated and is usually projected into the complex domain space. Therefore, Formula \eqref{Eq.3} can be described by the Procrustean distance \cite{small1996distributions}:
\begin{equation}
    d_{p}[O(v_{1}),O(v_{2})]=\cos^{-1}\left(\left|\sum_{j=1}^{p} v_{1j}v_{2j}^*\right|\right), \tag{4} \label{Eq.4}
\end{equation}
where $v_{1j}$ and $v_{2j}$ denote the $j$-th complex coordinates of $v_1$ and $v_2$, respectively, and $v_{2j}^{*}$ is denoted as the complex conjugate of $v_{2j}$. Subsequently, researchers found that data redundancy could be reduced based on the shape space theory for fast object recognition \cite{zhang1998invariant}. Further work includes calculating the Geodesic distance between two images using the corresponding feature points, thus calculating the degree of similarity of key objects in the two images \cite{han2013recognize}.

In the pre-shape space, points on a Geodesic curve can be used to describe a set of changing shapes. Evans et al. proposed a method to fit the data in the pre-shape space by obtaining a set of orthogonal points in the pre-shape space to construct the Geodesic curve \cite{evans2005curve,kenobi2010shape}. Based on this, Han et al. proposed a method to construct the Geodesic curve and thus expand the data in the pre-shape space using some small amount of data, where the two endpoints of the curves can be non-orthogonal \cite{han2014recognizing}. Assuming that $v_1$ and $v_2$ denote two points in the pre-shape space, the Geodesic curve formula can be derived as follows:
\begin{equation}
    \Gamma_{(v_1,v_2)}(s)= (\cos s)\cdot v_1 + (\sin s)\frac{v_2 - v_1 \cdot \cos \theta_{(v_1,v_2)}}{\sin \theta_{(v_1,v_2)}},(0\leq s \leq \theta_{(v_1,v_2)}) \tag{5} \label{Eq.5}
\end{equation}
where $\theta_{(v_1,v_2)}$ denotes the Geodesic distance between $v_1$ and $v_2$, and $s$ is the angle that denotes the Geodesic distance between the new point $v$ and $v_1$. If $0 \leq s \leq \theta_{(v_1,v_2)}$, then $v_{1} \leq \Gamma_{(v_1,v_2)} \leq v_{2}$. Following the curve, a series of new data or points can be generated in the pre-shape space, and all of these points are equivalent variations between $v_1$ and $v_2$, which is extremely helpful for expanding the size of the dataset. 

The shape space theory has been applied to various vision-related tasks, such as action recognition \cite{kumar2024survey}, object recognition \cite{han2014recognizing}, and point cloud registration \cite{lv2023kss}. Existing methods typically treat two pre-shapes as fixed endpoints to construct Geodesic curves for data augmentation. These generated samples encapsulate rich latent transformation information between two pre-shapes. In practical scenarios, however, multiple pre-shapes often coexist, rendering existing pairwise Geodesic construction methods inadequate for leveraging holistic dataset information. To address this limitation, we propose a more practical method for Geodesic curve construction among multiple pre-shapes.

\section{Methods}

\subsection{Image feature extraction}
The initial step of the FAGC involves mapping the input data from the pixel space to the pre-shape space. Typically, the quality of features is significantly influenced by the feature extraction method. Here, we adopt the deep learning method. Complex neural network models have a tendency to extract highly discriminative features, and the utilization of high-quality features can significantly improve the performance of downstream tasks. The image features can be extracted from each layer of the neural network models.  The higher layers of the network often exhibit a stronger capability to learn more intricate and comprehensive feature representations.
In this paper, we employ a Vision Transformer (ViT) network \cite{dosovitskiy2020image} of which the final fully connected layer is removed as the feature extractor.
The samples are fed into the pre-trained ViT network for fine-tuning, and the fine-tuning process is continued until the cross-entropy loss stabilizes.
The output features from the trained ViT network represent the features in the feature space.

\subsection{Projection of the image features into the pre-shape space}\label{3.2}
According to the theory of shape space, the vertices of a shape structure are typically required to be two-dimensional or three-dimensional. Therefore, it is necessary to expand the dimensions of the features extracted from the ViT network before projecting them into the pre-shape space or shape space. 
The up-dimensioned step can be achieved by adding additional dimensions or using techniques such as interpolation or transformation to match the required dimensional space of the shape structure.

In this paper, the up-dimensioned method is to copy a feature into two identical features. The image feature is assumed to be represented as $V_0=\{x_1,x_2,...,x_n \}$, where $n$ is the dimension of the feature. 
Thus, the up-dimensioned function $f(x)=x$, i.e., $V_0$ is transformed to $V_1=\{x_{1},y_{1},x_{2},y_{2},...,x_{n},y_{n} |y_{i}=x_{i},i\in [1, n]\}$.
To mitigate the effect of the feature positions, the means of the features are removed, and the new feature $V_2=\{x_{1}^{'},y_{1}^{'},x_{2}^{'},y_{2}^{'},...,x_{n}^{'},y_{n}^{'} | x_{i}^{'}=x_{i} - X_{\mu},y_{i}^{'}=y_{i}-Y_{\mu},i\in [1,n]\}$, where $X_\mu, Y_\mu$ are the means of $\{x_i\}$ and $\{y_i\}$, respectively.
Then, the scaling of the feature $V_2$ is excluded with formula $z = \frac{V_2}{||V_2||}$, where $||V_2||$ means the Euclidean norm of $V_2$.
Thus, the image feature $V_0$ is projected into a pre-shape space as feature vector $z$.

\subsection{Construction of the Geodesic curves to fit the feature vectors}
To construct a Geodesic curve in the pre-shape space, more than two image samples are required. In this paper, the number of Geodesic curves matches the number of categories, meaning that the construction of the Geodesic curves is performed within the feature vectors of the same category. The feature vectors within the same category exhibit smaller variations, making it easier to define the labels for newly augmented feature vectors. As mentioned in the above subsections, the features are extracted from the multiple images with the ViT network model. Then, the features are projected into the pre-shape space, denoted as vectors $Z=\{z_{1}^{1},...,z_{1}^{M_{1}},...,z_{N}^{1},...,z_{N}^{M_{N}}\}$, where $N$ represents the total number of image categories, $M_i$ represents the number of images in the $i$-th category, and $z_i^j$ denotes $j$-th sample of the $i$-th category.

\begin{figure*}[htbp]
    \centering
    \includegraphics[width=\textwidth]
    {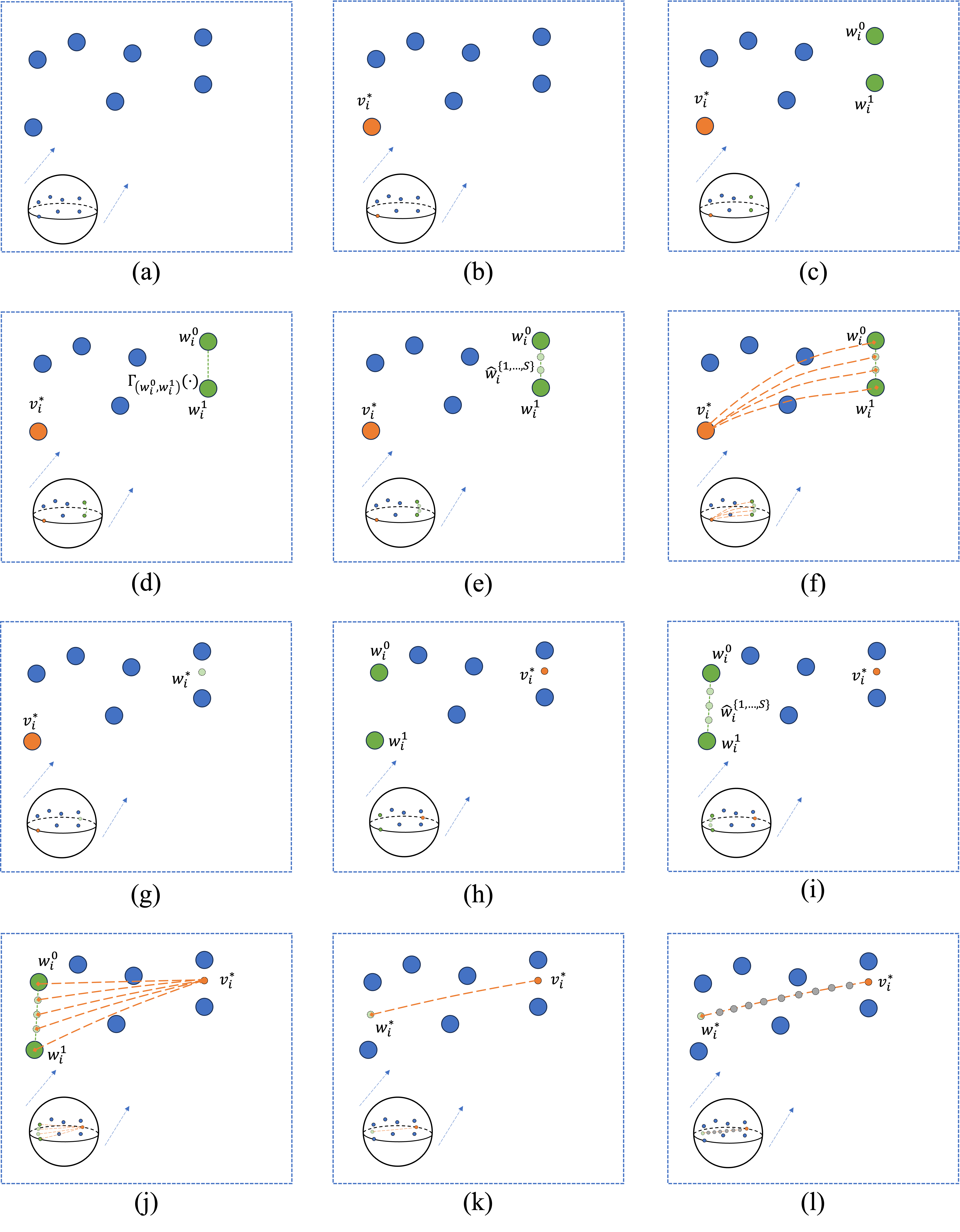}
    \caption{\rmfamily{The process of constructing an optimal Geodesic curve and generating new feature vectors using the FAGC. The hyper-sphere is a pre-shape space; the blue solid dots indicate the feature vectors in the pre-shape space; the solid orange dots represent feature vectors for parameter $v_{i}^{*}$; the solid green dots represent the two farthest known feature vectors $w_{i}^{0}$ and $w_{i}^{1}$ from feature vector $v_{i}^{*}$. (a) Projection of image features to the pre-shape space; (b) initialization of parameter $v_i^*$; from (c) to (g), the process of finding $w_{i}^{*}$; from (h) to (k), the process of finding new $v_i^*$; (l) obtaining an optimal Geodesic curve $\Gamma_{(v_i^*,w_i^*)}(\cdot)$ and augmented feature vectors.} }
    \label{fig:Geodesic_curve}
\end{figure*}

The set of the Geodesic curves corresponding to different image categories in the pre-shape space is denoted as $\{\Gamma_{(v_{i}^{*},w_{i}^{*})}|i\in (1,N)\}$, where $v_{i}^{*}$ and $w_{i}^{*}$ represents the endpoints of the optimal Geodesic curve, as $v_1$ and $v_2$ in Formula \eqref{Eq.5}. 
To obtain the  $v_{i}^{*},w_{i}^*$ of the optimal Geodesic curve $\Gamma_{(v_{i}^{*},w_{i}^{*})}$, the first step is to calculate the Geodesic distance among all feature vectors. Then, the feature vector with the maximum sum of distances from other feature vectors is selected as the initialization parameter $v_{i}^{*}$, defined as follows:

\begin{equation}
    v_{i}^{*} = \mathop{\mathrm{argmax}}_{z_{i}^{k}}\sum_{j=1}^{M_i}d(z_{i}^{k},z_{i}^{j}),k\in \{1,...,M_{i}\}
    \tag{6}
\end{equation}
where $d(z_{i}^{k},z_{i}^{j} )$ means the Geodesic distance between $z_{i}^{k}$ and $z_{i}^{j}$, calculated with Formula \eqref{Eq.2}.
Subsequently, $w_i^*$ is obtained through Geodesic curve $\Gamma_{(w_i^0,w_i^1)}(\cdot)$, which is determined by the two feature vectors $w_{i}^{0}$ and $w_{i}^{1}$ furthest from $v_{i}^{*}$.
As shown in Figure \ref{fig:Geodesic_curve} (c), the two feature vectors, $w_i^0$ and $w_i^1$, corresponding to the maximum two Geodesic distance from $v_i^*$, are calculated using the following formulas: 

\begin{align}
    &w_{i}^{0}=\mathop{\mathrm{argmax}}_{{Q_{i}-\{v_{i}^{*}\}}}
    \left(d\left(v_{i}^{*},{Q_{i}-\{v_{i}^{*}\}}\right)\right), and \tag{7} \notag \\
    &w_{i}^{1}=\mathop{\mathrm{argmax}}_{{Q_{i}-\{v_{i}^{*},w_{i}^{0}\}}}
    \left(d\left(v_{i}^{*},{Q_{i}-\{v_{i}^{*},w_{i}^{0}\}}\right)\right), \tag{8} \notag
\end{align}
where $Q_{i}=\{z_{i}^{1},z_{i}^{2},...,z_{i}^{M_i}\}$ to represent the set of feature vectors for the $i$-th category $\{z_{i}^{1},z_{i}^{2},...,z_{i}^{M_{i}}\}$. $Q_i-\{\#\}$ means if the feature vectors $\#$ are in $Q_i$, they are removed from $Q_i$.

Then, a Geodesic curve $\Gamma_{(w_i^0,w_i^1)}(\cdot)$ between $w_i^0$ and $w_i^1$ is calculated with Formula \eqref{Eq.5}. A set of feature vectors $\hat{w}_i^{\{1,...,S\}}$ is obtained by following the curve $\Gamma_{(w_i^0,w_i^1)}(\cdot)$, denoted as $\{\Gamma_{(w_{i}^{0},w_{i}^{1})}(s_{i}^{1}),\Gamma_{(w_{i}^{0},w_{i}^{1})}(s_{i}^{2}),...,\Gamma_{(w_{i}^{0},w_{i}^{1})}(s_{i}^{S})\}$, where $S$ represents the number of temporary feature vectors on $\Gamma_{(w_i^0,w_i^1)}(\cdot)$. Finally, among $v_i^*$ and $\hat{w}_i^{\{1,...,S\}}$, several Geodesic curves $\Gamma_{(v_i^*,\hat{w}_i^{\{1,...,S\}})}(\cdot)$ are described. The sum of squared Geodesic distance is calculated from the $i$-th category of feature vectors $z_i^1,z_i^2,...,z_i^{M_i}$ to  $\Gamma_{(v_i^*,\hat{w}_i^{\{1,...,S\}})}(\cdot)$. The feature vectors $\hat{w}_i^{*}$ corresponding to the shortest sum is calculated as follows:

\begin{equation}
    \hat{w}_{i}^{*} = \mathop{\mathrm{argmin}}_{\hat{w}_{i}^{k}}\sum_{j=1}^{M_i}
    \Bigl(d\left(z_{i}^{j},\Gamma_{(v_{i}^{*},\hat{w}_{i}^{k})}(\cdot)\right)\Bigr)^{2},k\in \{1,...,S\} \tag{9}
\end{equation}
and the step is illustrated in Figure \ref{fig:Geodesic_curve} (f) and Figure \ref{fig:Geodesic_curve} (g). 
Subsequently, assign the variable $w_i^*$ to $v_i^*$ and calculate the two furthest feature vectors from $v_i^*$, as illustrated in Figure \ref{fig:Geodesic_curve} (h). Repeat the steps for obtaining $w_i^*$ as previously described in Figures \ref{fig:Geodesic_curve} (i) to (j) until $w_i^*$ and $v_i^*$ cease to change. Consequently, $\Gamma_{(v_i^*,w_i^*)}(\cdot)$ denotes the optimal Geodesic curve fitting the distribution of feature vector $z_{i}^{*}$, as depicted in Figure \ref{fig:Geodesic_curve} (k). The detailed process of the FAGC processing specific category feature vectors is outlined in Algorithm ~\ref{alg1}.

All Geodesic curves for different categories, denoted as $\{\Gamma_{(v_i^*,w_i^*)}(\cdot) | i \in \{1,N\}\}$, can be constructed with the above mentioned method. And, with utilizing Formula \eqref{Eq.2}, the Geodesic distance $d(v_i^*,w_i^*)$ between all pairs of $v_i^*$ and $w_i^*$ is calculated. The distance indicates the range of the angle value $s$ within the Geodesic curve function $\Gamma_{(v_i^*,w_i^*)}(\cdot)$. A uniform distribution $s \sim U(0,d(v_i^*,w_i^*))$ is employed to sample a set of angle values $\{s_i^j | j \in \{1,K\}\}$, where $K$ represents the number of sampled angle values for the Geodesic curve of the $i$-th category.
The new feature vectors $Z'=\{z_1^{'1},...,z_1^{'K},...,z_N^{'1},...,z_N^{'K}\}$ are generated by substituting $s_i^j$ into the the Geodesic curve of the category $i$. These feature vectors and the original feature vectors from the sample image, constitute a set of augmented feature vectors $\{z_i^j, z_i^{'k} | i \in (1, N), j \in (1, M_i), k \in (1, K)\}$. 

\subsection{Design of the loss function}
To better balance the weight of original and augmented feature vectors during the training process, we optimized the loss function for the application of the FAGC. Generally, the cross entropy loss function is chosen since it can more sensitively reflect the differences between the predicted and true categories. In general, the number of augmented feature vectors with the FAGC is far more than the number of the original feature vectors. To balance the influence of the quantitative differences, a new loss function is proposed based on the cross entropy loss. We employ the standard cross entropy loss function, denoted as $L_r$, to evaluate the model's performance on original feature vectors. For the augmented feature vectors, two key control factors are introduced, i.e., the random probability factor $P_g$ and the influence factor $\lambda$. 
Let $\sigma$ be a random variable uniformly distributed in \([0,1]\), then the $P_{g}$ can be computed as
\begin{equation}
    P_{g} = \begin{cases} 
    1 & \text{if } \sigma > 0.5 \\
    0 & \text{otherwise}
\end{cases} .
\end{equation}
In each training batch, the final classifier loss function $L_{cls}$ is defined as follows:
\begin{equation}
    L_{cls}=L_{r}+P_g\times\lambda\times L_{g}. \tag{10} \label{Eq.10}
\end{equation}
If $P_{g} = 1$, the model is trained with the batch of the augmented feature vectors. Otherwise, $P_{g} = 0$ and the augmented feature vectors are ignored. $P_{g}$ can be used to randomly choose partially augmented feature vectors for the training process. $\lambda$ adjusts the contribution of the augmented feature vectors in the overall loss function and takes values between 0 and 1. Thus, an influence balance between the original feature vectors and the augmented feature vectors is controlled with $P_{g}$ and $\lambda$.

\begin{algorithm}
\caption{Feature Augmentation on the Geodesic Curve in the Pre-Shape Space (FAGC)}
\label{alg1}
\textbf{Input:} Feature vectors $Q_i=\{z_{i}^{1},...,z_{i}^{M_{i}}\}$ in the pre-shape space, number of samples in $i$-th category $M_i$, number of augmented feature vectors $K$,
and number of temporary feature vectors on the Geodesic curve $S$.\\
\textbf{Output:} The Geodesic curve $\Gamma_{(v_i^*,w_i^*)}(\cdot)$ and the augmented feature vectors $Z'=\{z_i^{'1},...,z_i^{'K}\}$.

\begin{algorithmic}[1]
    \STATE $v_i^* =\mathop{\arg\max}\limits_{z_i^{k}}\sum_{j=1}^{M_i}d(z_i^k,z_i^j),k\in\{1,...,M_i\}$;\vspace{1.5ex}
    \WHILE{$v^*,w^*$ not converge}\vspace{1.5ex}
        
        \STATE $w^0 = \mathop{\arg\max}\limits_{{Q_i-\{v_i^*\}}}\left(d(v_i^*,{Q_i-\{v_i^*\}})\right)$;\vspace{1.5ex}
        
        \STATE $w^1 = \mathop{\arg\max}\limits_{{Q_i-\{v_i^*,w^0\}}}\left(d(v_i^*,{Q_i-\{v_i^*,w^0\}})\right)$;\vspace{1.5ex}
        
        \STATE 
        $\hat{w}_i^{\{1,...,S\}}=\{\Gamma_{(w^0,w^1)}(s\cdot 1),$ \\
        $\qquad\Gamma_{(w^0,w^1)}(s\cdot2),...,\Gamma_{(w^0,w^1)}(s\cdot S)\}$;\vspace{1.5ex}

        \STATE $w_i^*=\mathop{\arg\min}\limits_{\hat{w}_i^k}\sum_{j=1}^{M_i}(d(z_i^j,\Gamma_{(v_i^*,\hat{w}_i^k)}(\cdot)))^2,$ \\
        $\qquad k\in\{1,...,S\}$;\vspace{1.5ex}

        \STATE Assign $w_i^*$ to $ v_i^*$;\vspace{1.5ex}

    \ENDWHILE\vspace{1.5ex}
    \STATE Calculate  $d(v_i^*,w_i^*)$ with Formula\eqref{Eq.2};\vspace{1.5ex}
    \FOR{$k=1$ to $K$}\vspace{1.5ex}
        \STATE Sample $\hat{s}_i^k \sim U(0,d(v_i^*,w_i^*))$;\vspace{1.5ex}
        \STATE $\hat{z}_i^k=\cos (\hat{s}_i^k)\cdot v_i^* +
        \sin(\hat{s}_i^k) \frac{w_i^*-v_i^*\cdot \cos\theta_{(v_i^*,w_i^*)}}{\sin\theta_{(v_i^*,w_i^*)}}$;\vspace{1.5ex}
    \ENDFOR
\end{algorithmic}
\end{algorithm}

\section{Experiments}
In this section, the FAGC is validated across several commonly small-sample scenarios, encompassing different learning frameworks and datasets of varying scales and difficulties.

\subsection{Experimental settings }
The FAGC framework operates in an offline manner and is structured into three distinct phases: feature extraction, feature augmentation, and downstream task integration. 
Since the performance of the FAGC depends on the quality of the features, the ViT  \cite{dosovitskiy2020image} series networks, primarily ViT-S \cite{dosovitskiy2020image} and ViT-T \cite{dosovitskiy2020image}, are used as feature extractors. Except for Section \ref{4.4}, the experimental data undergo a preprocessing process that includes horizontal flips and random crops before feature extraction.
The standard cross-entropy loss function is employed, with AdamW \cite{loshchilov2017decoupled} as the optimizer. Additional configurations include a weight decay of 0.0005, a learning rate of 0.0005, a layer decay rate of 0.5, and a moving average momentum of 0.9. The $\lambda$ is set to 0.5. All experiments are run on a server equipped with an NVIDIA 4090 GPU, and the reported results represent the average of three experimental runs. Other settings related to specific experimental content will be explained in subsequent sections.

In this paper, the FAGC is first integrated into the semi-supervised learning paradigm. Since real-world applications often face a scarcity of data labels, semi-supervised learning represents one of the most common approaches in deep learning. Unlike traditional machine learning pipelines, which is inherently compatible with the FAGC's offline training, applying the FAGC to deep learning requires certain adaptations.
Specially, when integrating the FAGC into a target deep model, the architecture of model is modularized into two core components: a feature extractor and a classifier. In practice, the model is first trained on the real data using conventional deep learning methods, and the trained model is retained. Subsequently, features are extracted from the input data and passed through the FAGC for augmentation. The augmented feature vectors, combined with the original feature vectors, are then utilized to fine-tune the classifier, thereby enhancing the model's overall performance.

\subsection{Applying the FAGC in existing semi-supervised or supervised methods}\label{4.2}
The FAGC framework is combined with various state-of-the-art semi-supervised or supervised learning methods on the CIFAR-10 and CIFAR-100 datasets \cite{krizhevsky2009learning}.
CIFAR-10 and CIFAR-100 are benchmark datasets comprising 50,000 training and 10,000 test images, which are in color and have a resolution of 32×32 pixels. CIFAR-100 shares the identical image set with CIFAR-10, consisting of 100 categories, with each category comprising 600 images.
The training set is partitioned into labeled and unlabeled subsets, respectively.
The numbers of samples per category for two experiments on the CIFAR-10 are 4 and 20, respectively, Thus, the total labeled samples are 40 and 200 for the two experiments, respectively. 
Similarly, for the CIFAR-100 datasets, we maintain the same sample ratios, yielding labeled samples of 400 and 2000 for the two scenarios. 
Considering the image size of the CIFAR dataset, the input scale of the standard ViT is adjusted to accommodate 32×32 image input and patch size of 2. We use ViT-T and ViT-S respectively to process CIFAR-10 and CIFAR-100.
The semi-supervised methods include CoMatch \cite{li2021comatch}, $\Pi$-Model \cite{rasmus2015semi}, Mean Teacher \cite{tarvainen2017mean}, Pseudo Labeling \cite{lee2013pseudo}, MixMatch \cite{berthelot2019mixmatch}, VAT \cite{miyato2018virtual}, Dash \cite{xu2021dash}, CRMatch \cite{fan2023revisiting}, UDA \cite{xie2020unsupervised}, FixMatch \cite{sohn2020fixmatch}, FlexMatch \cite{zhang2021flexmatch}, FreeMatch \cite{wang2022freematch}, AdaMatch \cite{berthelot2021adamatch}, SimMatch \cite{zheng2022simmatch}, SoftMatch \cite{chen2023softmatch}.

In our comparative analysis, the FAGC is integrated with different semi-supervised models of which source codes are referenced from USB \cite{wang2022usb}. 
For each model, we gather experimental results from the last 10 epochs which are used to calculate the average accuracy. Simultaneously, we report a 95\% confidence interval.
The results are shown in Table \ref{tab:FAGC-effectiveness}. Obviously, the FAGC can combine different models to enhance the performance of the original model. Particularly, for the 400 labeled images in CIFAR-100, the FAGC achieves an average classification accuracy improvement of approximately 4\%. For the labeled images in CIFAR-10, the FAGC exhibits a significant improvement of about 1.5\% when utilizing 40 labeled images. However, the enhancement of the FAGC is not significant if the number of training samples is large, such as 200 labeled images in CIFAR-10. This is because the backbone network is more thoroughly trained with 400 images, and its accuracy reaches 90.13\%.

To describe the influence of the FAGC combined with different backbone networks, the FAGC is combined with ViT-T, ViT-S, WRN-28-2 \cite{zagoruyko2016wide}, and WRN-28-8 \cite{zagoruyko2016wide} in supervised training. 
As shown in Table \ref{tab:FAGC-ful_sup}, the performance of the models are significantly improved by the FAGC. With the increase of the training samples, the influence of the FAGC is decreased. Therefore, the FAGC delivers its maximum effectiveness under conditions of extreme data scarcity.

\begin{table*}[!ht]
\caption{The impact of the FAGC on the classification accuracies of different semi-supervised models. ``AVG'' (the average accuracy of all methods) is used to demonstrate the universal improvement effect achieved by adding the FAGC across different methods.}
\renewcommand{\arraystretch}{1.2}
\resizebox{\textwidth}{!}{
\begin{tabular}{ccccccccc}
\toprule
& \multicolumn{4}{c}{CIFAR-100} & \multicolumn{4}{c}{CIFAR-10}\\
\midrule
\multicolumn{1}{c}{\multirow{2}{*}{Methods}} & 
\multicolumn{2}{c}{400 labels} &
\multicolumn{2}{c}{2000 labels} &
\multicolumn{2}{c}{40 labels} &
\multicolumn{2}{c}{200 labels} \\
\cmidrule{2-9}
& - & \textbf{+FAGC} & - & \textbf{+FAGC} & - & \textbf{+FAGC} & - & \textbf{+FAGC} \\
\midrule
CoMatch\cite{li2021comatch} & 62.37±0.12 & 68.10±0.21 & 80.71±0.13 & 82.84±0.24 & 88.24±0.05 & 90.95±0.17 & 93.64±0.11 & 94.39±0.21 \\
$\Pi$-Model\cite{rasmus2015semi} & 54.04±0.23 & 69.29±0.34 & 82.90±0.16 & 84.12±0.19 & 80.66±0.13 & 82.56±0.22 & 92.90±0.14 & 92.77±0.37 \\
Mean Teacher\cite{tarvainen2017mean} & 70.50±0.22 & 75.96±0.14 & 81.95±0.11 & 83.68±0.26 & 81.21±0.17 & 84.33±0.19 & 92.64±0.03 & 92.43±0.25 \\
Pseudo Labeling\cite{lee2013pseudo} & 73.51±0.25 & 76.29±0.22 & 82.61±0.15 & 84.42±0.21 & 81.92±0.21 & 82.44±0.26 & 94.15±0.12 & 94.57±0.23 \\
MixMatch\cite{berthelot2019mixmatch} & 73.87±0.32 & 76.06±0.15 & 84.07±0.21 & 84.58±0.34 & 81.81±0.25 & 85.07±0.19 & 93.83±0.16 & 94.31±0.31 \\
VAT\cite{miyato2018virtual} & 77.33±0.17 & 80.40±0.32 & 83.91±0.07 & 85.43±0.31 & 89.92±0.04 & 90.21±0.11 & 95.28±0.05 & 95.14±0.36 \\
Dash\cite{xu2021dash} & 78.74±0.23 & 82.69±0.23 & 86.46±0.03 & 87.17±0.16 & 93.20±0.05 & 93.55±0.21 & 95.30±0.07 & 95.90±0.25 \\
CRMatch\cite{fan2023revisiting} & 79.84±0.15 & 80.54±0.26 & 86.12±0.17 & 87.16±0.22 & 92.79±0.08 & 93.31±0.32 & 95.41±0.13 & 95.89±0.29 \\
UDA\cite{xie2020unsupervised} & 80.51±0.14 & 82.52±0.31 & 86.31±0.15 & 87.50±0.18 & 91.74±0.11 & 92.13±0.25 & 95.33±0.12 & 95.78±0.22 \\
FixMatch\cite{sohn2020fixmatch} & 80.59±0.25 & 82.96±0.23 & 86.76±0.04 & 87.46±0.36 & 91.40±0.12 & 91.84±0.23 & 95.08±0.09 & 95.35±0.35 \\
FlexMatch\cite{zhang2021flexmatch} & 80.78±0.15 & 82.86±0.25 & 86.12±0.15 & 87.16±0.19 & 92.57±0.21 & 93.08±0.15 & 95.75±0.05 & 95.63±0.27 \\
AdaMatch\cite{berthelot2021adamatch} & 81.93±0.17 & 83.78±0.22 & 85.84±0.23 & 87.13±0.32 & 90.95±0.12 & 92.96±0.23 & 95.12±0.03 & 95.58±0.32\\
FreeMatch\cite{wang2022freematch} & 82.22±0.13 & 83.34±0.25 & 86.27±0.23 & 87.56±0.28 & 93.57±0.15 & 94.07±0.24 & 95.73±0.04 & 96.11±0.12 \\
SimMatch\cite{zheng2022simmatch} & 82.48±0.12 & 83.57±0.34 & 86.98±0.06 & 87.88±0.37 & 92.78±0.18 & 93.18±0.16 & 95.49±0.11 & 95.77±0.24 \\
SoftMatch\cite{chen2023softmatch} & 82.72±0.11 & 83.11±0.28 & 86.83±0.17 & 87.81±0.26 & 93.27±0.11 & 93.68±0.12 & 95.72±0.16 & 95.55±0.34 \\
\midrule
AVG & 75.88±0.17 & \textbf{79.18}±0.25 & 84.69±0.13 & \textbf{85.96}±0.26 & 88.45±0.13 & \textbf{89.76}±0.21 & 94.46±0.09 & \textbf{94.81}±0.27 \\
\bottomrule
\end{tabular}\label{tab:FAGC-effectiveness}
}
\end{table*}

\begin{table*}[!ht]
\caption{The impact of the FAGC on the classification accuracies of different backbone models.}
\renewcommand{\arraystretch}{1.2}
\resizebox{\textwidth}{!}{
\begin{tabular}{ccccccccccccc}
\toprule
& \multicolumn{6}{c}{CIFAR-10} & \multicolumn{6}{c}{CIFAR-100}\\
\midrule
\multirow{2}{*}{Methods} &
\multicolumn{2}{c}{40 labels} &
\multicolumn{2}{c}{200 labels} &
\multicolumn{2}{c}{1000 labels} &
\multicolumn{2}{c}{400 labels} &
\multicolumn{2}{c}{2000 labels} &
\multicolumn{2}{c}{10000 labels} \\
\cmidrule{2-13}
& - & \textbf{+FAGC} & - & \textbf{+FAGC} & - & \textbf{+FAGC} & - & \textbf{+FAGC}
& - & \textbf{+FAGC} & - & \textbf{+FAGC} \\
\midrule
ViT-T & 79.32 & 82.83 & 90.13 & 91.77 & 96.21 & 96.35 & 63.02 & 66.69 & 74.39 & 76.54 & 80.96 & 82.61 \\
ViT-S & 90.00 & 93.31 & 96.45 & 97.12 & 98.19 & 98.26 & 72.65 & 75.36 & 81.20 & 83.43 & 86.28 & 87.39 \\
WRN-28-2 & 92.89 & 93.77 & 93.92 & 94.17 & 94.30 & 94.58 & 68.92 & 72.09 & 74.73 & 76.33 & 75.59 & 76.21 \\
WRN-28-8 & 93.29 & 94.51 & 93.32 & 94.53 & 95.32 & 95.68 & 75.73 & 77.21 & 78.82 & 79.56 & 81.23 & 82.12 \\
\bottomrule
\end{tabular}
\label{tab:FAGC-ful_sup}
}
\end{table*}

\begin{table}[!ht]
\centering
\caption{Comparison of the FAGC with other feature augmentation methods on different datasets. ``STL'', ``ANI'' and ``CUB'' represent datasets STL-10, Animals-90, and CUB-200-2011, respectively.}
\setlength{\tabcolsep}{1.5mm}{
\begin{tabular}{cccc}
\toprule
Methods & STL & ANI & CUB \\
\midrule
Baseline & 94.73±0.05 & 96.40±0.09 & 67.49±0.31 \\
Mani-Mix-up \cite{verma2019manifold} & 94.39±0.08 & 96.60±0.15 & 68.17±0.25 \\
SFA-S \cite{li2021simple} & 94.96±0.05 & \textbf{97.20±0.16} & 68.85±0.19\\
MoEx \cite{li2021feature} & 94.44±0.18 & 96.60±0.07 & 67.95±0.21\\
FL-SMOTE \cite{liu2024feature} & 94.60±0.12 & 96.80±0.08 & 68.17±0.23\\
FAGC & \textbf{95.80±0.15} & 97.00±0.07 & \textbf{69.30±0.28}\\
\bottomrule
\end{tabular}
}
\label{tab:multi}
\end{table}

\subsection{Comparison with other feature augmentation methods on additional datasets}
To further validate the effectiveness and applicability of the FAGC, comparisons with other feature augmentation methods are conducted on several more challenging datasets. The datasets include STL-10 (STL) \cite{coates2011analysis}, CUB-200-2011 (CUB) \cite{WahCUB_200_2011}, and Animals-90 (ANI) \cite{banerjee_animals90}. STL-10 contains 10 categories of data, with 500 images for training and 800 images for testing in each category. CUB-200-2011 contains a total of 11788 images belonging to 200 categories of birds, of which 5994 are used for training and the rest for testing. Animals-90 is a collection of 5400 animal images from 90 categories collected by google images. The feature augmentation methods used for comparison include Mani-Mix-up \cite{verma2019manifold}, SFA-S \cite{li2021simple}, MoEx \cite{li2021feature}, and FL-SMOTE \cite{liu2024feature}.

We extracted a subset from each dataset for training to demonstrate the effectiveness of the FAGC in extremely small-sample scenarios. The subset consists of 10 categories, with only 4 labeled samples per category for training. The feature extraction network used is ViT-S, and the classifier is the MLP. The results are shown in Table \ref{tab:multi}, the addition of the FAGC further improved the classification accuracy when facing more challenging data. Regardless of the initial accuracy, the improvement of model performance by the FAGC demonstrates its applicability. When compared with various feature augmentation methods, the FAGC achieved the most significant performance improvement in the majority of cases.

\subsection{The combination of the FAGC and common pixel-level augmentation methods.} \label{4.4}
As a feature-level augmentation method, the FAGC can be combined with existing pixel-level augmentation methods.
We choose several representative pixel-level augmentation methods which include horizontal flip, rotate, random crop, color jitter, cutout and augmix \cite{hendrycks2019augmix}. 
The experiments deploy ViT-S network on the CIFAR-10 dataset with 400 labeled data. 
Table ~\ref{table3} describes the comparison of performance before and after combining these methods with the FAGC. 
From Table ~\ref{table3}, the FAGC can improve the classification accuracy combined with other pixel-level augmentation methods.
It is noteworthy that the combination of all pixel-level augmentation methods does not achieve optimal classification accuracy, as shown in ID 8 of Table ~\ref{table3}. In contrast, optimal accuracy is achieved with the combination of the methods of horizontal flip, random crop and the FAGC.

\begin{table*}[!ht]
\caption{\rmfamily{Comparison of the compatibility between the FAGC and various pixel-level augmentation methods.}}
\renewcommand{\arraystretch}{1.2}
\normalsize
\centering
\resizebox{\textwidth}{!}{
\begin{tabular}{c|cccccc|cc}
\toprule
\multirow{2}{*}{ID} & \multicolumn{6}{c|}{Pixel-level augmentation methods} & \multicolumn{2}{c}{Methods}\\
\cmidrule{2-9}
& horizontal flip & rotate & random crop & color jitter & cutout & augmix & - & \textbf{+FAGC} \\
\midrule
1 & \checkmark & \checkmark & & & & & 70.23 & 72.34 \\
2 & & & \checkmark & & & & 70.21 & 73.39 \\
3 & & & & \checkmark & & & 71.12 & 72.13 \\
4 & & & & & \checkmark & & 72.14 & 74.93 \\
5 & \checkmark & & \checkmark & & & & 72.65 & 75.36 \\
6 & & \checkmark & & & \checkmark & & 70.11 & 72.28 \\
7 & \checkmark & & \checkmark & & & \checkmark & 72.48 & 73.54 \\
8 & \checkmark & \checkmark & \checkmark & \checkmark & \checkmark & \checkmark & 66.38 & 68.27\\
\bottomrule
\end{tabular}}
\label{table3}
\end{table*}

\subsection{Different machine learning classifiers with the FAGC}
The experiments are conducted by selecting 40 and 200 labeled images from CIFAR-10 as the training set to investigate the impact of the FAGC on different feature classifiers.
The feature extraction network model is ViT-S, and feature classifiers for comparison include KNN \cite{cover1967nearest}, SVC \cite{cortes1995support}, DecisionTree \cite{quinlan1986induction}, ExtraTree \cite{geurts2006extremely}, RandomForest \cite{breiman2001random}, Bagging \cite{breiman1996bagging}, GradientBoost \cite{friedman2001greedy}, and MLP \cite{rumelhart1986learning}. 
The results indicate that the addition of the FAGC can improve the performance of these machine learning classifiers by approximately 4$\%$, as shown in Table \ref{table4}. Meanwhile, it can be observed that the FAGC has varying impacts on different classifiers. The FAGC has the greatest improvement on MLP, as MLP loads pre-trained parameters from the ViT-S output layer.

\begin{table*}[!ht]
\begin{center}
\normalsize
\caption{The impact of the different feature classifiers on the performance with the FAGC on CIFAR-10.}
\resizebox{\textwidth}{!}{
\begin{tabularx}{\textwidth}{c*{4}{>{\centering\arraybackslash}X}}
\toprule
& \multicolumn{4}{c}{CIFAR-10}\\
\midrule
\multirow{2}{*}{Classifiers} &
\multicolumn{2}{c}{40 labels} &
\multicolumn{2}{c}{200 labels} \\
\cmidrule{2-5}
& - & \textbf{+FAGC} & - & \textbf{+FAGC} \\
\midrule
KNN\cite{cover1967nearest} & 79.54 & 91.11 & 96.29 & 96.75 \\
SVC\cite{cortes1995support} & 89.02 & 91.29 & 96.72 & 96.85 \\
DecisionTree\cite{quinlan1986induction} & 56.76 & 58.51 & 85.62 & 86.03 \\
ExtraTree\cite{geurts2006extremely} & 51.37 & 66.07 & 78.68 & 85.89 \\
RandomForest\cite{breiman2001random} & 91.36 & 91.23 & 96.09 & 96.37 \\
Bagging\cite{breiman1996bagging} & 74.52 & 77.87 & 94.38 & 94.52 \\
GradientBoost\cite{friedman2001greedy} & 71.32 & 76.41 & 88.48 & 89.81 \\
MLP\cite{rumelhart1986learning} & 90.00 & \textbf{93.31} & 96.45 & \textbf{97.12} \\
\bottomrule
\end{tabularx}
}
\label{table4}
\end{center}
\end{table*}

\subsection{The different number of the augmented feature vectors} \label{Sec:ablation-study}
\begin{figure}[htbp]
    \centering
    \includegraphics[width=0.65\textwidth]{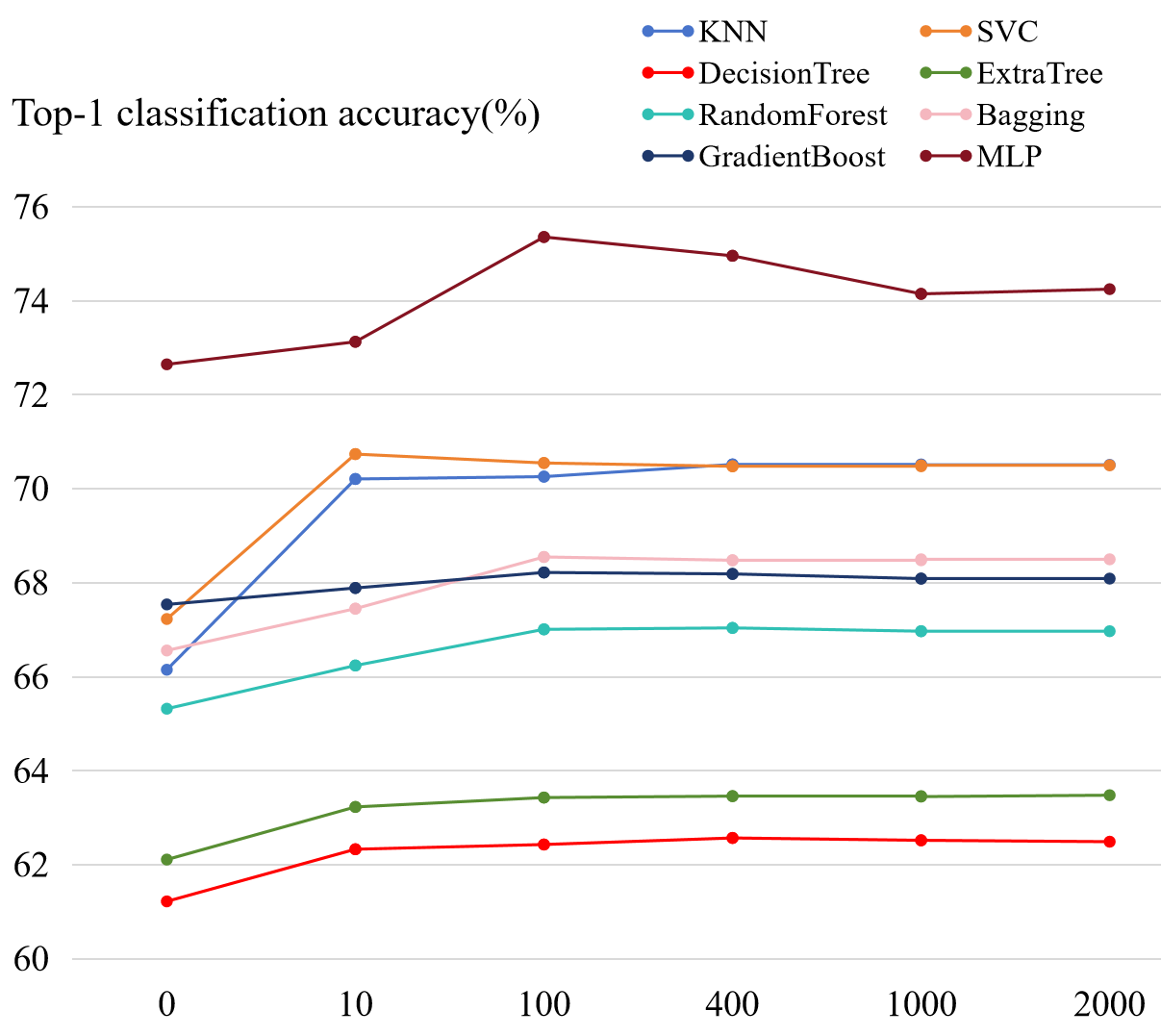}
    \caption{The impact of the number of the augmented feature vectors on model performance under different feature classifiers. The X-axis represents the number of the augmented feature vectors, and the Y-axis represents the classification accuracy with different classifiers.}
    \label{fig:ml_gen_fea}
\end{figure}

\begin{figure}[htbp]
    \centering
    \includegraphics[width=0.65\textwidth]{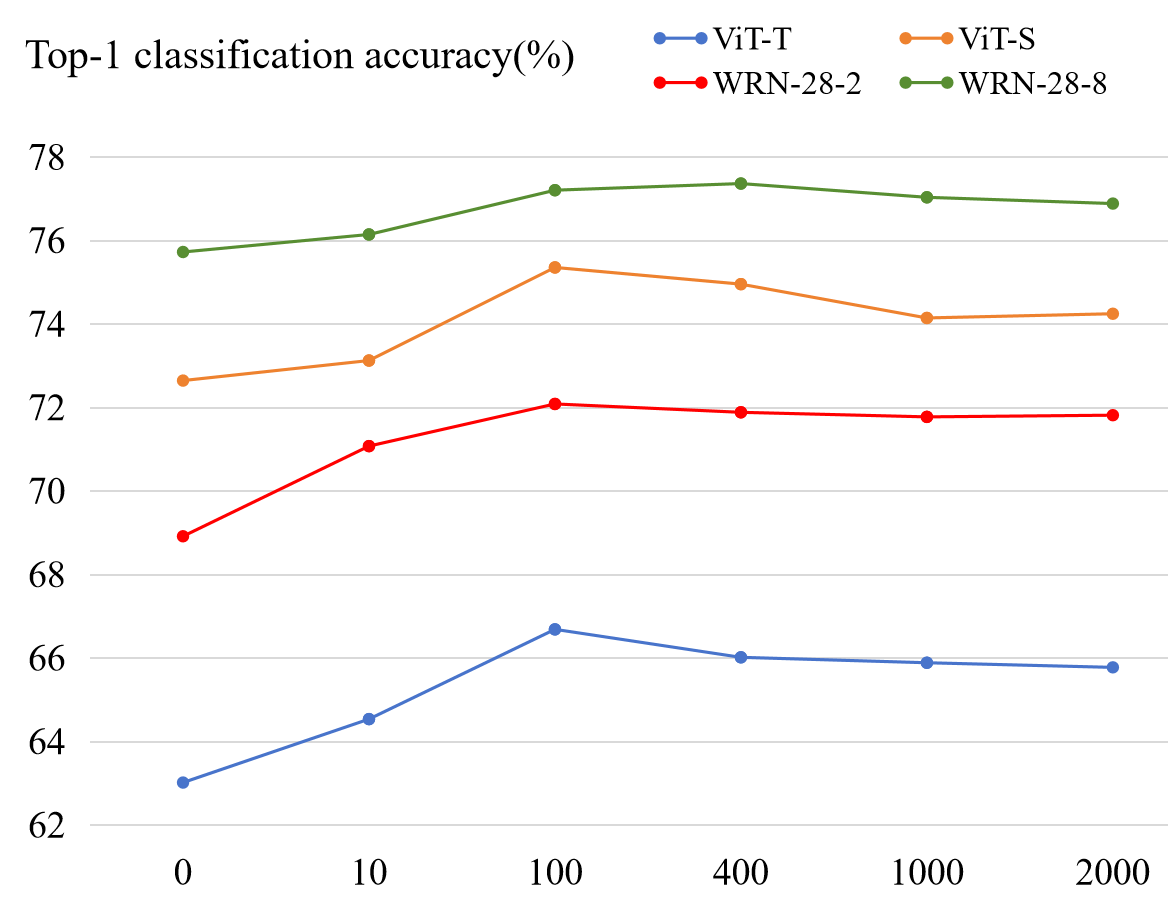}
    \caption{The impact of the number of augmented features on model performance under different feature extraction models. The X-axis represents the number of the augmented feature vectors, and the Y-axis represents the classification accuracy with different feature extraction models.}
    \label{fig:mlp_num_gen_fea}
\end{figure}

To evaluate the influence of the number of augmented feature vectors for the models, we use multiple machine learning models as classifiers trained on 400 labeled features in CIFAR-100, including KNN, SVC, DecisionTree, ExtraTree, RandomForest, Bagging, GradientBoost and MLP. The feature extraction network model is ViT-S, and the numbers of the augmented feature vectors are 10, 100, 400, 1000, and 2000. The results of the experiments are presented in Figure \ref{fig:ml_gen_fea}. During the process of increasing the number from 10 to 2000, the performance of all classifiers exhibited a consistent pattern: rapid initial improvement, followed by a gradual decline, and eventual stabilization.
The number of the generated feature vectors does not exhibit a strict correlation with the model's classification accuracy. Excessive augmented feature vectors can cause the model to overly focus on their distribution, but ignore the distribution of real feature vectors. Moreover, the best performance is achieved when using the MLP classifier with the number of augmented feature vectors set to 100. Supplementary experiments are conducted utilizing different feature extraction models, including ViT-T, ViT-S, WRN-28-8, and WRN-28-2. The number of the augmented feature vectors is also increased from 10 to 2000, while the classifier is consistently maintained as the MLP to observe the impact of different feature extraction models on the FAGC. The detailed experimental results are shown in Figure \ref{fig:mlp_num_gen_fea}. The results indicate that the performance improvement of the FAGC varies across different feature extraction models, with the most significant enhancement observed for ViT-T. As the number of the augmented feature vectors increases, the gain brought by the FAGC aligns with the conclusions from Figure \ref{fig:ml_gen_fea}, achieving optimal effectiveness when the number is set to 100.

\subsection{The influence of factor $\lambda$}
To evaluate the influence of the augmented features from the Geodesic curves on the performance of models, the experiments are conducted for the 40 labeled images on the CIFAR-10. ViT-T and ViT-S are selected as feature extraction models, and MLP is employed as the final classifier. The ablation experiments are conducted on the parameter $ \lambda$ in the classification loss, as the Formula \eqref{Eq.10}. 
As the value of $ \lambda$ increases, the model assigns greater weight to the knowledge derived from the augmented feature vectors.
The experimental results are shown in Table ~\ref{table5}. 
As the influence factor $\lambda$ increases, the classification accuracy of the model initially rises and then exhibits a slight decline. 
The declining trend is attributed to the increasing influence of the augmented features on the loss function, causing the model to ignore the real features.

\begin{table*}[htbp]
\caption{\rmfamily{The influence of factor $\lambda$ on the classification performance on CIFAR-10.}}
\scriptsize
\rmfamily
\begin{tabularx}{\linewidth}{>{\centering\arraybackslash}X *{9}{>{\centering\arraybackslash}X}}
\toprule
\multirow{2}{*}{Methods} & \multicolumn{9}{c}{$\lambda$} \\
\cmidrule{2-10}
& 0 & 0.1 & 0.3 & 0.45 & 0.5 & 0.55 & 0.7 & 0.9 & 1 \\
\midrule
ViT-T & 79.32 & 80.95 & 81.86 & 82.92 & 82.83 & 82.85 & 82.63 & 82.46 & 82.32 \\
ViT-S & 90.00 & 92.34 & 93.56 & 93.68 & 93.77 & 93.79 & 93.54 & 93.23 & 93.11 \\
\bottomrule
\end{tabularx}
\label{table5}
\end{table*}

\begin{figure}[htbp]
    \centering
    \includegraphics[scale=0.6]{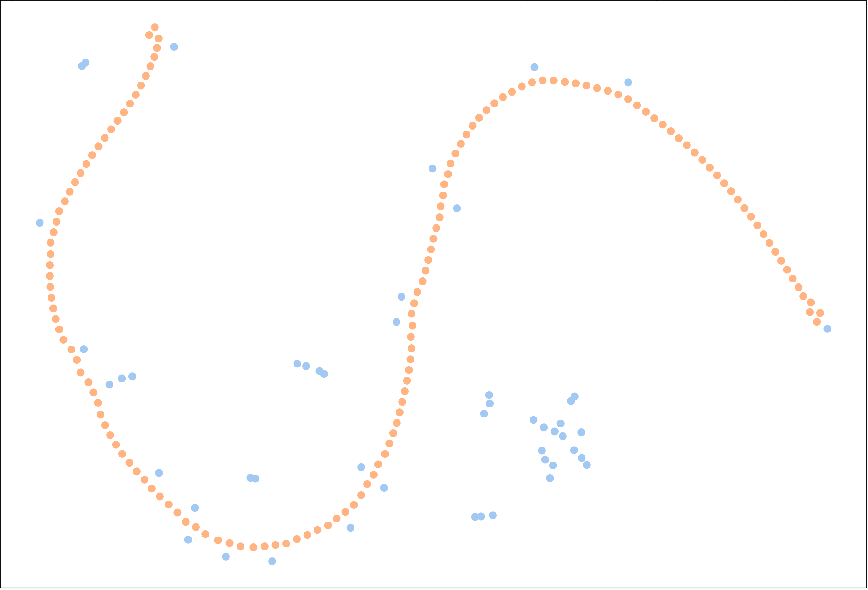}
    \caption{\rmfamily{The t-SNE visualization of the constructed Geodesic curve. The blue dots are the initial feature vectors, and the yellow dots are the augmented feature vectors sampled on the Geodesic curve.}}
    \label{figure6}
\end{figure}

\begin{figure*}[htbp]
    \centering
    \includegraphics[scale=0.7]{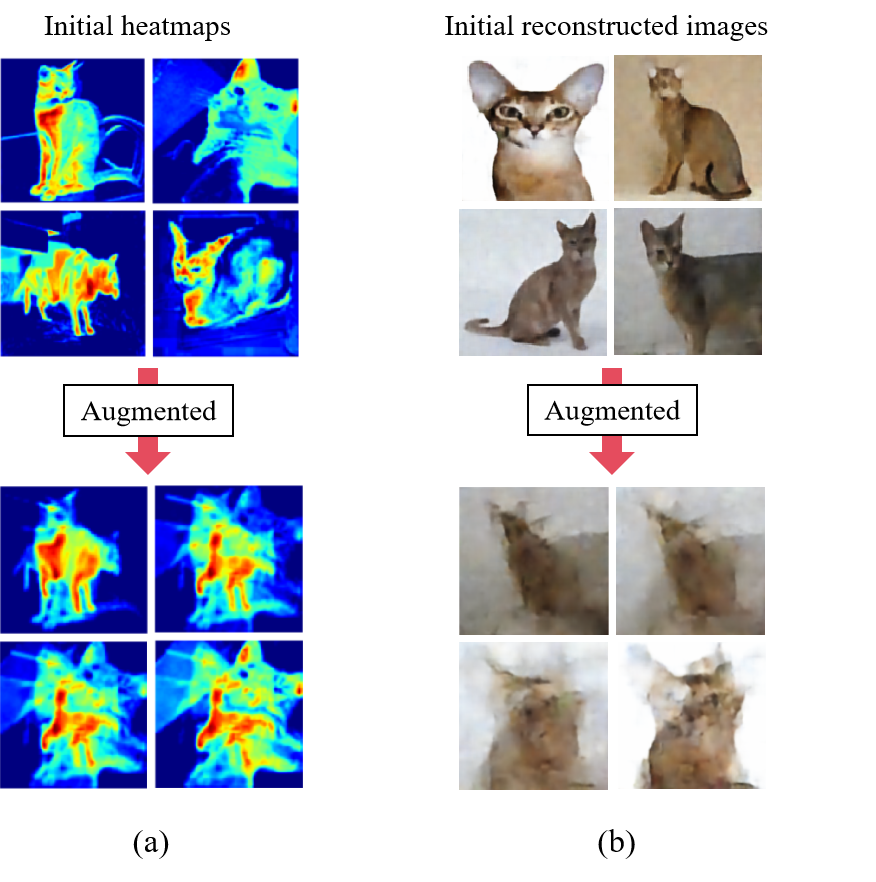}
    \caption{\rmfamily{Visualizations of the augmented feature vectors with the FAGC. (a) is the heatmaps generated directly from the feature vectors, and (b) is the reconstructed images obtained by decoding latent vectors using a pre-trained autoencoder.}}
    \label{figure7}
\end{figure*}

\subsection{Visualization}
For a clearer comparison between augmented and initial feature vectors, t-SNE, heatmaps, and autoencoder (AE) are used to visualize the augmented feature vectors. Specifically, the t-SNE technique visualizes feature vectors extracted from the CIFAR dataset by a pre-trained ResNet-18. In Figure \ref{figure6}, there are 50 initial feature vectors and 150 augmented feature vectors. It can be observed that the constructed Geodesic curve, i.e., the path formed by the yellow dots, fits the initial feature vectors well and shows no sign of overfitting. Furthermore, Figure \ref{figure7} reveals the semantic relationships between the augmented feature vectors and the initial feature vectors. The heatmaps in \ref{figure7}(a) are generated from the feature vectors of randomly selected cat images. 
In \ref{figure7}(b), the initial images are originated from the Oxford-IIIT Pets \cite{parkhi2012cats} dataset. The reconstructed images are generated by encoding the initial images using the pre-trained encoder, augmenting the feature vectors through the FAGC, and then reconstructing the images via the decoder. As demonstrated in Figure \ref{figure7}, the feature vectors augmented by the FAGC maintain semantic consistency with the original feature vectors, and their variation remains within controlled bounds. The points along the Geodesic curve, i.e., the augmented feature vectors, capture rich geometric transformation information between initial feature vectors with high confidence. Collectively, these visualization experiments validate the safety and reliability of the feature vectors augmented by the FAGC, demonstrating their suitability for many practical scenarios.

\section{Conclusion} \label{Sec:conclution}
This paper proposes a feature augmentation method in the pre-shape space, called the FAGC, which provides additional information for machine learning and deep learning models.
The FAGC encompasses the entire workflow of the feature extraction, shape modeling, Geodesic curve construction, and downstream tasks application, enabling its convenient deployment in diverse small-sample scenarios. Crucially, The Geodesic curve construction is the core innovation of the FAGC. Unlike existing approaches that simply connect pairs of data in the pre-shape space, the FAGC proposes a novel method to construct the Geodesic curves across multiple initial pre-shapes. The FAGC has been validated across numerous designed small-sample scenarios, with results demonstrating the effectiveness of the features augmented by the FAGC. Furthermore, visualization results across multiple scenarios further verify the safety and reliability of the FAGC.

The FAGC has successfully bridged the shape space theory with deep learning and machine learning, opening up new research avenues for data augmentation. However, the FAGC still has some areas for improvement. For instance, the Geodesic curves may not be the optimal geometric structure for data augmentation, and there is a need to explore diversified applications of the FAGC in deep learning models. In the future, we will explore more complex geometric structures in the pre-shape space, such as Geodesic surfaces, and investigate modular applications of the FAGC in deep learning frameworks.
%

\section {acknowledgements}
This research is sponsored by National Natural Science Foundation of China (Grant No. 52273228), Key Program of Science and Technology of Yunnan Province (202302AB080022), the Project of Key Laboratory of Silicate Cultural Relics Conservation (Shanghai University), Ministry of Education (No. SCRC2023ZZ07TS).

\bibliographystyle{FAGC}
\bibliography{FAGC}

\end{CJK}
\end{document}